\documentclass[letterpaper, 10 pt, journal, twoside]{ieeetran}
\usepackage[caption=false,font=normalsize,labelfont=sf,textfont=sf]{subfig}

\usepackage{times}
\usepackage{xcolor}
\usepackage{cite}
\usepackage{amssymb}
\usepackage[table]{xcolor}
\usepackage{array}

\usepackage{multicol}
\usepackage[bookmarks=true,
  pdftitle={Colosseum V2: Benchmarking Generalization for Vision-Language-Action Models},
  pdfauthor={Jeremy Morgan, Hyeonho Oh, Prajwal Vijay, Jincen Song, Ashvin Arora, Hojung Lim, Alina Du, Gaurav S. Sukhatme, Jesse Thomason, Ishika Singh},
  pdfsubject={IEEE Robotics and Automation Letters},
  pdfkeywords={Imitation Learning; Dual Arm Manipulation; Deep Learning in Grasping and Manipulation; Vision-Language-Action Models}
]{hyperref}
\usepackage{tikz}
\usepackage{array} 
\usepackage{graphicx}  
\usepackage{tabularx} 
\usepackage{amsmath}
\usepackage{booktabs}

\usepackage{microtype}

\usepackage{pifont}
\usepackage{colortbl}
\usepackage{cleveref}
\usepackage{xspace}

\usetikzlibrary{patterns,calc}

\newcommand{\piO}{\ensuremath{\pi_{0.5}}}
\newcommand{\act}{ACT}

\newcommand{\website}{\url{https://colosseum-v2.github.io/}}

\newcommand{\DualArmStackTwoCubes}{\textsc{DualArmStackTwoCubes}}

\newcommand{\LiftPegUpright}{\textsc{LiftPegUpright}}

\newcommand{\PickDishFromRack}{\textsc{PickDishFromRack}}
\newcommand{\PickSodaFromCabinet}{\textsc{PickSodaFromCabinet}}

\newcommand{\RaiseCube}{\textsc{RaiseCube}}
\newcommand{\RotateArrow}{\textsc{RotateArrow}}
\newcommand{\ScoopBanana}{\textsc{ScoopBanana}}

\newcommand{\numperturbationstotal}{19}

\newcommand{\bimanual}{Bimanual}
\newcommand{\singlearm}{Single-Arm}

\newcommand{\none}{\texttt{None}}
\newcommand{\mocolor}{\texttt{MO-Color}}
\newcommand{\motexture}{\texttt{MO-Texture}}
\newcommand{\rocolor}{\texttt{RO-Color}}
\newcommand{\rotexture}{\texttt{RO-Texture}}
\newcommand{\tablecolor}{\texttt{Table-Color}}
\newcommand{\tabletexture}{\texttt{Table-Texture}}
\newcommand{\lightcolor}{\texttt{Light-Color}}
\newcommand{\distractorobjects}{\texttt{Distractor-Objects}}
\newcommand{\backgroundcolor}{\texttt{Background-Color}}
\newcommand{\backgroundtexture}{\texttt{Background-Texture}}
\newcommand{\camerapose}{\texttt{Camera-Pose}}
\newcommand{\poserandomization}{\texttt{Pose-Randomization}}
\newcommand{\mosize}{\texttt{MO-Size}}
\newcommand{\rosize}{\texttt{RO-Size}}
\newcommand{\languagenone}{\texttt{Language-None}}
\newcommand{\languageparaphrase}{\texttt{Language-Paraphrase}}
\newcommand{\languagerandom}{\texttt{Language-Random}}
\newcommand{\languageothertask}{\texttt{Language-Other-Task}}

\newcommand{\allperturbations}{\texttt{All}}

\newcommand{\benchmark}{\textsc{Colosseum V2}}
\newcommand{\colosseum}{\textsc{Colosseum}}

\newcommand{\topfigureslidedown}{\vspace*{0pt}}

\begin{document}

\title{Colosseum V2: Benchmarking Generalization for Vision-Language-Action Models}

\author{
Jeremy Morgan$^{1}$,
Hyeonho Oh$^{1}$,
Prajwal Vijay$^{2}$,
Jincen Song$^{3}$,
Ashvin Arora$^{1}$,
Hojung Lim$^{1}$,
Alina Du$^{1}$,
Jesse Thomason$^{1}$,
Gaurav S. Sukhatme$^{1}$,
and Ishika Singh$^{1}$%
\thanks{Manuscript received: May 9, 2026; Revised: August 6, 2026; Accepted: August 30, 2026.}%
\thanks{This paper was recommended for publication by Editor Lucia Pallottino upon evaluation of the Associate Editor and Reviewers' comments.}%
\thanks{$^{1}$J. Morgan, H. Oh, A. Arora, H. Lim, A. Du, G. S. Sukhatme, J. Thomason, and I. Singh are with the Department of Computer Science, University of Southern California, Los Angeles, CA, USA. J. Morgan is the corresponding author ({\tt\footnotesize morganje@usc.edu}).}%
\thanks{$^{2}$P. Vijay is with the Department of Electrical Engineering, Indian Institute of Technology Madras, Chennai, India.}%
\thanks{$^{3}$J. Song is with the Fu Foundation School of Engineering and Applied Science, Columbia University, New York, NY, USA.}%
\thanks{Digital Object Identifier (DOI): see top of this page.}%
}

\maketitle


\begin{abstract}
Vision–Language–Action (VLA) models demonstrate promising generalization in robotic manipulation, driven by advances in large-scale vision and language pre-training. 
This progress can be misleading. 
Despite the zero-shot perception and language capabilities of VLAs, their overall task performance often degrades under distribution shifts, revealing gaps in how these systems translate high-level understanding into robust behavior. 
To systematically study this gap, we introduce \benchmark{}, a large-scale simulation benchmark for evaluating VLA generalization in robot learning across diverse conditions. 
The benchmark comprises 28 tasks spanning 13 task categories and two robot morphologies, covering a wide range of manipulation primitives and long-horizon behaviors. 
Built on the ManiSkill simulator, \benchmark{} enables fast, GPU-parallelized evaluation and supports both in-domain and out-of-domain testing at scale.
We evaluate state-of-the-art methods, including Action Chunking Transformers (ACT) and \piO{}, and reveal limitations in both base performance and generalization. 
We demonstrate strong correlations between simulation and real-world metrics that support the ecological validity of the benchmark.
By standardizing tasks, metrics, and evaluation protocols within a unified benchmark, \benchmark{} enables reproducible and fair comparisons, reduced evaluation overhead, and accelerated progress toward general-purpose robot policies. Complete results and supplementary videos are available on the project’s website: \website{}.

\end{abstract}


\section{Introduction}
\IEEEPARstart{V}{ision}-Language models (VLMs), such as ChatGPT~\cite{openai2022chatgpt} and SAM 2~\cite{ravi2024sam2}, have demonstrated remarkable generalization across diverse visual scenes and linguistic instructions.

These models benefit from large-scale pre-training on Internet-scale data, enabling them to encode novel inputs and understand previously unseen scenarios. 

Recent VLA models have shown impressive capabilities in robotic manipulation. 
By leveraging vision and language pre-training, these models can interpret natural language instructions and recognize  objects previously unseen in robotics training data.
However, unlike VLMs, their generalization remains limited when evaluated across the full spectrum of manipulation tasks encountered in real-world environments.

\begin{figure}[t]
    \centering
    \includegraphics[width=1.0\columnwidth]{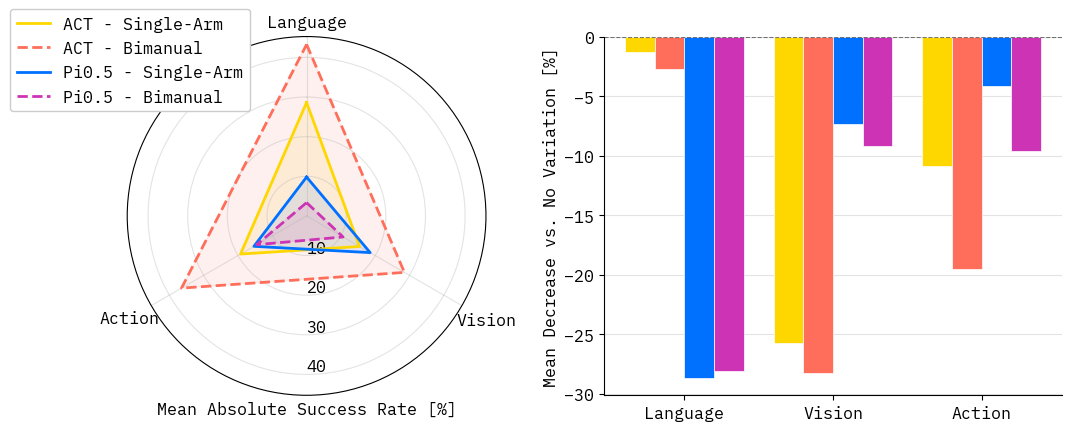}
    \caption{Comparison of overall generalization performance. The left plot shows average success rate across the vision, language, and action perturbations, whereas the right plot shows the average decrease in success rate compared to the no-perturbation condition. The \act{} models are largely unaffected by language perturbations but exhibit worse visual generalization rates compared to \piO{}.
    }
    \label{fig:radial_teasor}
\end{figure}

A key challenge lies in the mismatch between how different components of these models scale. 
While vision and language representations generalize broadly due to large and diverse pre-training, the downstream behavior of VLA systems is still shaped by comparatively limited robot interaction data.
As a result, improvements in perception and reasoning do not always translate into reliable task execution under novel conditions, leading to a gap between apparent and true generalization. 
A VLA may correctly interpret a new instruction or recognize a new object, yet still fail when the environment, phrasing, or interaction dynamics deviate from the robot data training distribution as shown in our overall generalization result in Figure~\ref{fig:radial_teasor}.
Such failures highlight that evaluating VLA models requires a comprehensive view of generalization that simultaneously considers perturbations in visual inputs, language instructions, and interaction dynamics.

%
%

\begin{figure*}[!ht]
    \topfigureslidedown
    \centering
    \includegraphics[width=1.0\textwidth]{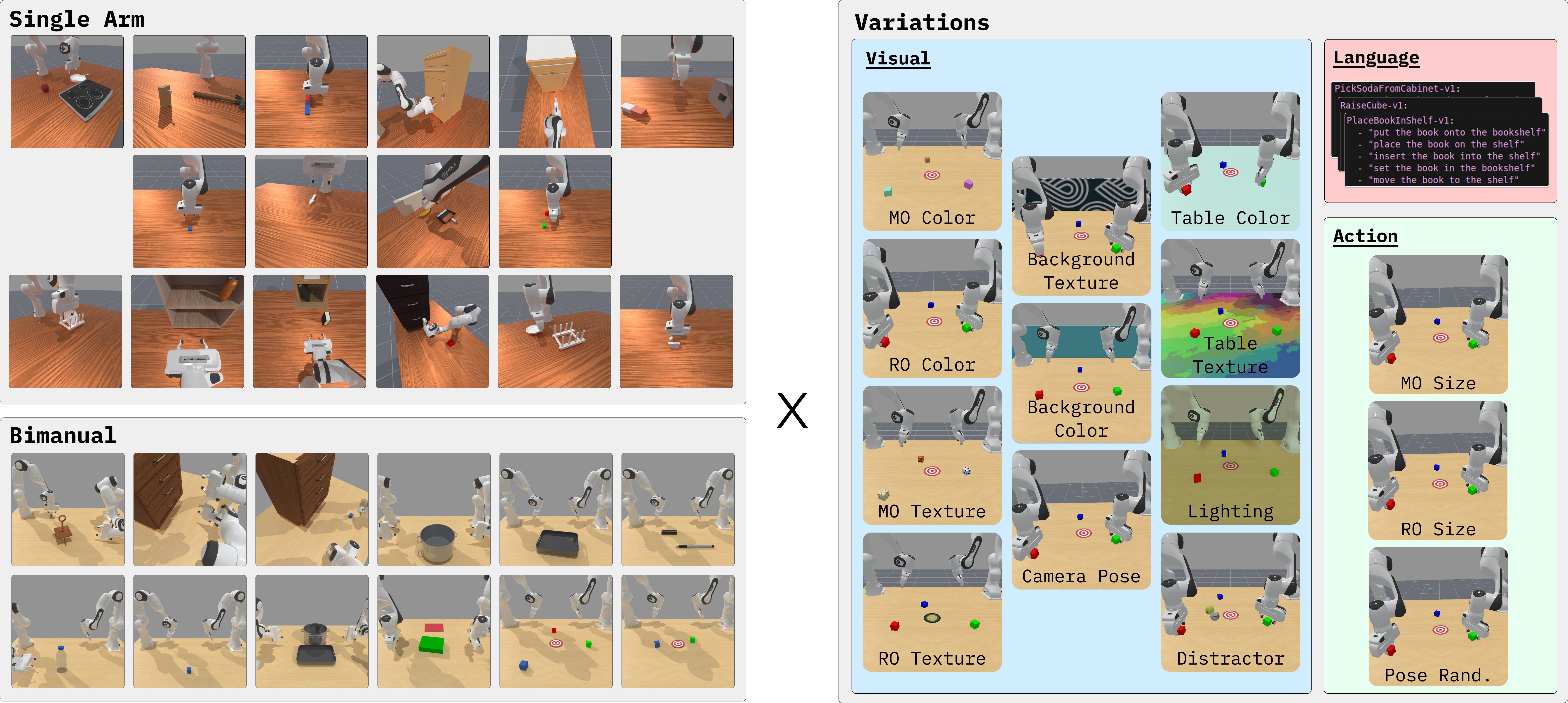}

    \caption{
        Overview of \benchmark{}. 
        Left: the full set of tasks across two robot morphologies (\singlearm{} and \bimanual{}), spanning diverse manipulation primitives and long-horizon behaviors. 
        Right: the perturbations used to evaluate visual, language, and action generalization. 
        In total, the benchmark comprises 28 tasks across 13 task categories with \numperturbationstotal{} controlled perturbation factors.
    }
    \label{fig:taskzoo}
\end{figure*}

Despite its importance, this gap remains difficult to measure using existing benchmarks. 
Many current evaluation suites focus on isolated aspects such as perception or language grounding, while manipulation benchmarks often evaluate a limited set of tasks or perturbations. 

To address this challenge, we introduce \benchmark{}, a large-scale simulation benchmark designed to systematically evaluate generalization across visual, language, and action axes (Figure~\ref{fig:taskzoo}). 
The benchmark comprises 28 tasks across 13 task categories and two robot morphologies, providing a diverse and structured evaluation protocol. 
Built on the ManiSkill simulator, \benchmark{} enables fast GPU-parallelized evaluation at scale.

We evaluate two state-of-the-art methods within this benchmark, including Action Chunking Transformers (ACT) and \piO. 
Our results (Figure~\ref{fig:radial_teasor}) reveal clear gaps between strong perception-driven capabilities and overall task robustness, highlighting key challenges that remain for deploying VLA systems in real-world settings.

In summary, this work includes the following contributions:

\begin{itemize}
    \item We introduce \benchmark{}, a large-scale benchmark specifically designed to measure the Visual, Language, and Action generalization of VLAs. The benchmark provides a comprehensive and structured evaluation protocol, enabling systematic and reproducible assessment of generalization.
    \item We leverage GPU-parallelized simulation to enable fast, large-scale evaluation, allowing statistically thorough benchmarking (e.g., 200 episodes per task-perturbation pair) to be completed in under half a day (Figure~\ref{fig:fps_comparison}). 



    \item We establish baselines by training and evaluating two state-of-the-art multitask Imitation Learning models, \act{} and \piO{}, and identify limitations in their generalization behavior and in-distribution performance.    
    
    \item We validate sim-to-real transfer by evaluating policies on real hardware, demonstrating that performance trends in simulation reflect real-world generalization behavior (Figure \ref{fig:hardware_vs_sim}).
\end{itemize}

\begin{table*}[t]
\centering
{\footnotesize

\begin{tabular}{p{0.15\textwidth} p{0.23\textwidth} p{0.28\textwidth} p{0.27\textwidth}}
\hline
\textbf{Dimension} &
\textbf{Limitation of V1} &
\textbf{Design choice in V2} &
\textbf{New capability / scientific question} \\
\hline

Evaluation scale &
Sequential, CPU-bound RLBench simulation limits the number of evaluation episodes. &
GPU-parallelized ManiSkill simulation with 200 episodes per task--perturbation pair. &
Enables statistically robust, large-scale comparisons across tasks, perturbations, and models. \\

Embodiment &
A single-arm Franka setup. &
Single-arm and bimanual Franka test suites with different action spaces. &
How does generalization change with embodiment, coordination requirements, and action-space dimensionality? \\

Task coverage &
20 tasks organized primarily by trajectory horizon. &
28 tasks across 13 categories, including bimanual and long-horizon behaviors. &
Do model weaknesses persist across manipulation primitives, task horizons, and morphologies? \\

Model scope &
Evaluation primarily focused on keyframe-based behavior-cloning policies. &
Evaluation of action-chunking and pretrained VLA models, including ACT and $\pi_{0.5}$. &
How do pretraining and model design affect base task performance and out-of-distribution robustness? \\

Language generalization &
Semantic language variation was not isolated as a controlled evaluation axis. &
Semantically equivalent instruction paraphrases are evaluated independently. &
Does vision--language pretraining produce robust language-conditioned behavior after robot-policy fine-tuning? \\

Generalization analysis &
Environmental perturbations were evaluated primarily as individual robustness factors. &
Perturbations are structured along visual, language, and action axes and evaluated across both morphologies. &
Does strong in-domain performance predict generalization, and which axes remain bottlenecks for modern VLAs? \\

\hline
\end{tabular}
}

\caption{Comparison of COLOSSEUM V1 and COLOSSEUM V2. V2 retains the original goal of evaluating robustness under distribution shift while enabling new analyses of modern VLA policies.}
\label{table:colosseum_v1_v2}
\end{table*}

%


\newcommand{\cmark}{\ding{51}}
\newcommand{\xmark}{\ding{55}}

\begin{figure*}[t]
\centering
\resizebox{\textwidth}{!}{

\footnotesize
\begin{tabular}{l|ccccccccc}
\toprule
Benchmark 
& Single Arm 
& Bimanual  
& GPU Sim 
& Visual Var. 
& Language Var. 
& Action Var. 
& Diverse Tasks \\
\midrule

RLBench
& \cmark & \xmark  & \xmark & \xmark & \cmark & \xmark & \cmark \\

Colosseum 
& \cmark & \xmark  & \xmark & \cmark & \cmark & \cmark & \cmark \\

LIBERO 
& \cmark & \xmark  & \xmark & \cmark & \cmark & \xmark & \cmark \\

Meta-World 
& \cmark & \xmark & \xmark & \xmark & \xmark & \xmark & \cmark \\

RoboSuite (Dexmimicgen/Robocasa)
& \cmark & \cmark & \xmark & \cmark & \xmark & \xmark & \cmark \\

RoboVerse 
& \cmark & \xmark & \cmark & \cmark & \xmark & \xmark & \cmark \\

ManiSkill 
& \cmark & \cmark & \cmark & \xmark & \xmark & \xmark & \cmark \\

CALVIN 
& \cmark & \xmark & \xmark & \xmark & \cmark & \xmark & \cmark \\

VLABench 
& \cmark & \xmark & \xmark & \cmark & \cmark & \cmark & \cmark \\

VLMbench 
& \cmark & \xmark & \xmark & \cmark & \xmark & \cmark & \cmark \\



\rowcolor{gray!10}
\textbf{\benchmark}
& \cmark & \cmark & \cmark & \cmark & \cmark & \cmark & \cmark \\
\bottomrule
\end{tabular}}

\caption{
Comparison of existing robot learning benchmarks and simulation platforms.
\benchmark{} supports diverse manipulation tasks, multiple embodiments, and GPU-parallelized evaluation through ManiSkill, enabling scalable benchmarking of modern robot policies. 
}
\label{table:benchmark_features}
\end{figure*}


\section{Background}

This section reviews robotics benchmarks and imitation learning methods, highlighting limitations in scalability, diversity, and generalization. It also discusses Vision–Language–Action models, noting their promise but continued reliance on large datasets and sensitivity to environment changes.


\subsection{Robotics Benchmarks}

\begin{figure}[t]
\topfigureslidedown
    \centering
    \includegraphics[width=0.98\columnwidth]{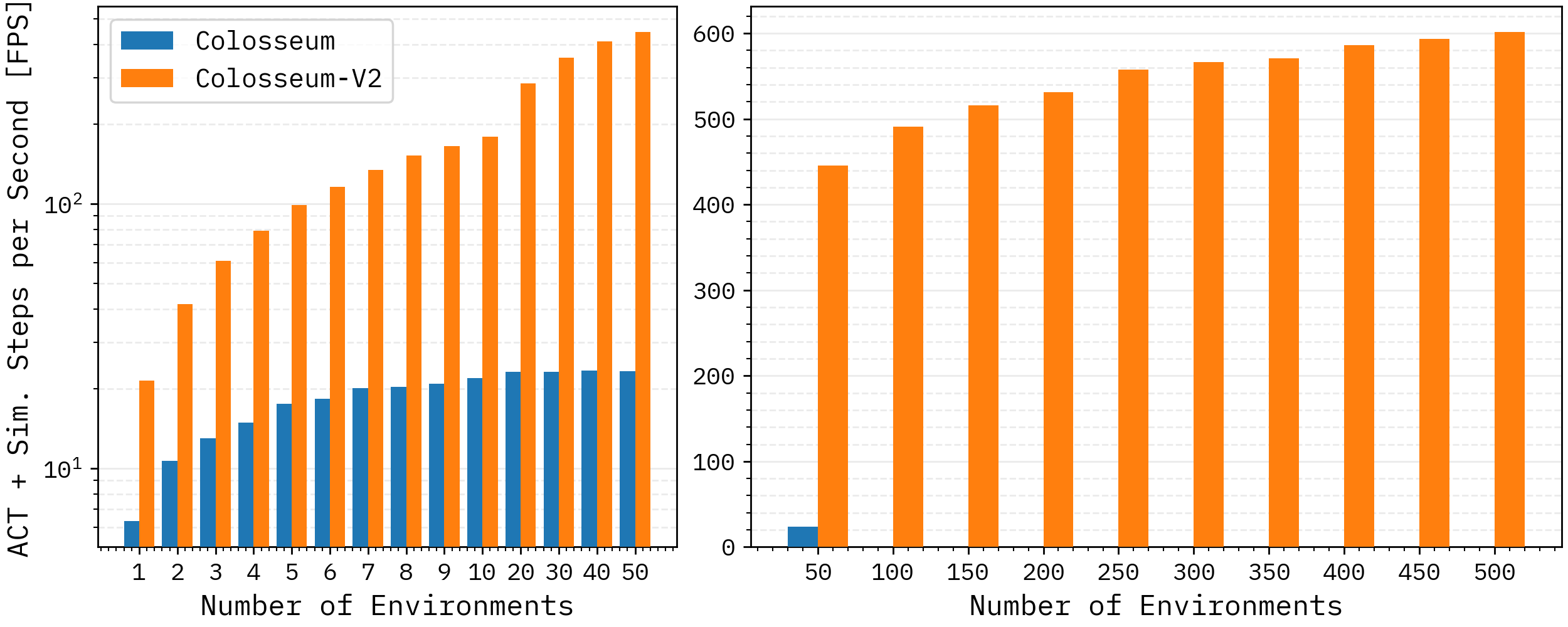}
    \caption{
    Simulator comparison. Frames per second (FPS) is computed as the number of parallel environments divided by the wall-clock time per simulation step (including \act{} inference). The left panel compares both simulators over the range supported by the original COLOSSEUM benchmark (up to 50 parallel environments), while the right panel extends the evaluation to 5000 parallel environments to illustrate the scalability of the GPU-parallel implementation.
    }
    \label{fig:fps_comparison}
\end{figure}

Large-scale benchmarks have played an important role in enabling systematic evaluation of robot learning methods. 
For example, simulation-based benchmark RLBench~\cite{james2019rlbenchrobotlearningbenchmark}, built on the PyRep interface~\cite{james2019pyrepbringingvrepdeep} and the CoppeliaSim simulator~\cite{coppeliaSim}, provides a standardized suite of 100 manipulation tasks designed for reinforcement learning and imitation learning research. 
While widely adopted, its underlying CPU-based simulator does not support vectorized execution and runs slower than real time, limiting scalability for large-scale training and evaluation.
Figure \ref{fig:fps_comparison} quantifies this bottleneck in addition to the gains achieved through GPU-parallelization.

The \colosseum\ benchmark~\cite{pumacay2024colosseum} builds on RLBench and extends it with structured generalization tests designed to evaluate the robustness of imitation learning models to visual and physical perturbations.
A comparison to related benchmarks~\cite{james2020rlbench,pumacay2024colosseum,liu2024libero,zhu2020robosuite,murali2020roboverse,taomaniskill3,mees2022calvin,zhang2024vlabenchlargescalebenchmarklanguageconditioned,zheng2022vlmbenchcompositionalbenchmarkvisionandlanguage} is provided in Figure~\ref{table:benchmark_features} while a detailed comparison to the Colosseum is shown in Table \ref{table:colosseum_v1_v2}.



Other benchmarks have been proposed to evaluate diverse aspects of robot learning. LIBERO~\cite{liu2024libero} focuses on lifelong learning, introducing task suites that disentangle declarative and procedural knowledge during sequential training. LIBERO-Para~\cite{kim2026liberoparadiagnosticbenchmarkmetrics} extends LIBERO with language paraphrasing to measure robustness to language instructions. RoboVerse~\cite{murali2020roboverse} provides a lightweight simulation platform for large-scale reinforcement learning experiments. Platforms such as RoboArena~\cite{roboarena} and RobotWin~\cite{robotwin} aim to evaluate generalist robot policies in distributed or large-scale environments. Bimanual Manipulation Benchmark~\cite{bimanual} focuses on bimanual manipulation in the absence of test time changes in the environment.
CALVIN~\cite{mees2022calvin} studies long-horizon, language-conditioned manipulation with compositional skill chaining. VLABench~\cite{zhang2024vlabenchlargescalebenchmarklanguageconditioned} and VLMbench~\cite{zheng2022vlmbenchcompositionalbenchmarkvisionandlanguage} focus on vision-language reasoning and compositional generalization, with less emphasis on control robustness. ManipBench~\cite{zhao2025manipbenchbenchmarkingvisionlanguagemodels} evaluates the reasoning capabilities of VLAs through multiple choice questions.






Despite these advances, most existing benchmarks either focus on narrow task families or evaluate policies under limited environmental perturbations. In contrast, \benchmark\ is designed to evaluate robustness and generalization across diverse manipulation tasks and environment perturbations.

\subsection{Imitation Learning for Robotic Manipulation}

Robotic manipulation has seen significant progress through learning-based approaches in both simulation and real-world settings. 
To improve generalization, pretrained visual representations such as R3M~\cite{r3m}, MVP~\cite{mvp}, and VIP~\cite{ma2022vip} provide transferable features for downstream robotic control tasks. Other works leverage pretrained vision-language models to ground manipulation policies in semantic understanding~\cite{cliport,huang2023voxposer}. 

Manipulation systems have shifted from direct continuous control to predicting \emph{keyframes} or \emph{action chunks}~\cite{c2farm,sundaresan2023kite,act-aloha}. 3D action representations have proven particularly effective for data-efficient learning. For example, PerAct~\cite{shridhar2022peract}, RVT~\cite{goyal2023rvt,goyal2024rvt}, and Act3D~\cite{gervet2023act3d} learn voxel-based action predictions conditioned on language instructions and visual observations. These approaches demonstrate strong performance with smaller numbers of demonstrations relative to standard BC, often requiring only tens of demonstrations per task. However, such methods are typically trained from scratch for specific robots and environments, limiting their ability to generalize to new objects, scenes, or instructions.

\textbf{Vision--Language--Action Models (VLA)}
models aim to combine large-scale pretrained vision and language representations with robot control policies. By leveraging semantic priors learned from internet-scale data, these models enable robots to interpret open-ended language instructions and generalize to previously unseen concepts~\cite{driess2023palm,brohan2023rt2}.
Several recent works demonstrate the promise of this paradigm. PaLM-E~\cite{driess2023palm} integrates large language models with robot perception and control, enabling reasoning over multimodal inputs. RT-2~\cite{brohan2023rt2} extends robotic transformers with internet-scale vision-language pretraining to enable semantic generalization. OpenVLA~\cite{kim24openvla} trains a unified vision-language-action policy on large-scale robot datasets and demonstrates strong cross-task transfer. More recent systems such as $\pi_0$~\cite{black2024pi0}, $\pi_{0\text{-fast}}$~\cite{pertsch2025pi0fast}, and $\pi_{0.5}$~\cite{intelligence2025pi05visionlanguageactionmodelopenworld} further explore scaling VLA models to broader environments and robot embodiments.

Despite these advances, current VLA approaches require extremely large demonstration datasets. For example, ${\text{OpenVLA}}$~\cite{kim24openvla} is trained on more than 900k demonstrations from the Open-X Embodiment dataset~\cite{padalkar2023open}. Even at this scale, such models remain sensitive to perturbations in camera viewpoints, scene geometry, and object configurations, as we show in our results (Section~\ref{sec:results}).

\begin{table}[t]
    \centering
    {\footnotesize
    \begin{tabularx}{\columnwidth}{
        >{\hsize=1.1\hsize}X
        >{\hsize=0.75\hsize}X
        >{\hsize=1.15\hsize}X
    }
        \toprule
        Task Name & Morphology & Task Category \\
        \midrule
        
        
        RaiseCube & \singlearm{} PJG & Pick \\
        PickSodaFromCabinet & \singlearm{} PJG & Pick \\
        PickDishFromRack & \singlearm{} PJG & Pick \\
        
        StackCube & \singlearm{} PJG & Pick \& Place \\
        PlaceBookInShelf & \singlearm{} PJG & Pick \& Place \\
        PlaceDishInRack & \singlearm{} PJG & Pick \& Place \\
        
        LiftPegUpright & \singlearm{} PJG & Reorientation \\
        RotateArrow & \singlearm{} PJG & Reorientation \\
        
        PegInsertionSide & \singlearm{} PJG & Insertion \\
        PlugCharger & \singlearm{} PJG & Insertion \\
        
        HammerNail & \singlearm{} PJG & Tool Use \\
        ScoopBanana & \singlearm{} PJG & Tool Use \\
        
        OpenDrawer & \singlearm{} PJG & Articulated  \\
        OpenCabinet & \singlearm{} PJG & Articulated \\
        
        PlaceCubeInDrawer & \singlearm{} PJG & Long-Horizon \\
        CookItemInPan & \singlearm{} PJG & Long-Horizon \\

        \midrule

        DualArmCubeHandover & \bimanual{} PJG & Handover \\
        DualArmBottleHandover & \bimanual{} PJG & Handover \\
        
        DualArmLiftPot & \bimanual{} PJG & Lift \\
        DualArmLiftTray & \bimanual{} PJG & Lift \\
        
        DualArmPushBox & \bimanual{} PJG & Reorientation \\
        DualArmPourPot & \bimanual{} PJG & Reorientation \\
        
        DualArmThreading & \bimanual{} PJG & Insertion \\
        DualArmPenCap & \bimanual{} PJG & Insertion \\

        DualArmDrawerPlace & \bimanual{} PJG & Articulated \\
        DualArmDrawerOpen & \bimanual{} PJG & Articulated \\
        DualArmStackCube & \bimanual{} PJG & Long-Horizon \\
        DualArmStackTwoCubes & \bimanual{} PJG & Long-Horizon \\

        \bottomrule
    \end{tabularx}
    }
    \caption{Tasks in \benchmark{}, grouped by robot morphology and task category.}
    
    \label{tab:pjg_tasks}
\end{table}

\section{Benchmark Design}

The benchmark comprises 28 simulation tasks spanning two robot setups: (1) a \singlearm{} manipulator equipped with a parallel-jaw gripper (PJG), and (2) a \bimanual{} configuration with two PJG-equipped manipulators. In both cases, the underlying robot model is the Franka Panda. The \bimanual{} setup places two Panda arms on opposite sides of a shared workspace, enabling coordinated dual-arm manipulation. We refer to the two setups as distinct robot morphologies. Each morphology includes 6–7 task categories designed to cover a broad spectrum of manipulation behaviors. The complete set of tasks and task categories is presented in Table \ref{tab:pjg_tasks}.

For the \singlearm{} tasks, the task categories are: Pick, Pick \& Place, Reorientation, Insertion, Tool Use, Articulated Object Interaction, and Long-Horizon Manipulation. The \bimanual{} tasks follow a closely related taxonomy, adapted to account for dual-arm coordination and shared-object manipulation.




\subsection{Task Categorization}
\label{method-tc}

The task categories are designed to represent a diverse and progressively challenging set of manipulation behaviors. Within each morphology, categories are ordered from relatively simple primitives to complex, temporally extended tasks. 

\subsubsection{\singlearm{} PJG} 

\textbf{Pick.}
This category involves grasping and lifting objects using the gripper. Task difficulty increases from grasping simple objects (e.g. a cube) to partially constrained items, such as extracting a plate from a drying rack.

\textbf{Pick \& Place.}
These tasks involve grasping, transporting, and placing an object within a target region. Representative examples include stacking blocks and placing objects on shelves.

\textbf{Reorientation.}
Reorientation tasks involve modifying an object’s pose to match a task-specific orientation.  Typical behaviors to reach the target orientation include rotating, flipping, or twisting objects.

\textbf{Insertion.}
This category emphasizes precision manipulation under tight geometric constraints. Representative tasks include peg-in-hole insertions and connector assembly, where success depends on precise alignment between objects and is achieved through controlled contact.

\textbf{Tool Use.}
Tool-use tasks require manipulating one object to effect change in another. Examples include hammering, scooping, stirring, or poking, all of which introduce indirect interaction dynamics.

\textbf{Articulated Object Interaction.}
These tasks involve interacting with articulated mechanisms such as drawers, cabinets, doors, or laptops. Successful execution requires reasoning about joint types (e.g., revolute or prismatic), respecting motion constraints, and applying sustained forces.

\textbf{Long-Horizon Manipulation.}
Long-horizon tasks compose multiple primitive actions during a trajectory. 


\subsubsection{\bimanual{} PJG}

\textbf{Handover.}
Handover tasks require transferring an object between two end-effectors, testing inter-arm coordination and timing. 

\textbf{Lift.}
In this category, both grippers cooperatively lift and transport a shared object (e.g., tray, box, pot). The primary challenge lies in maintaining balance and consistent force distribution while coordinating dual contact points.

\textbf{Reorientation.}
These tasks require coordinated manipulation to modify an object’s orientation using both arms simultaneously. 
 
\textbf{Insertion.}
This category extends \singlearm{} insertion tasks to the \bimanual{} setting. Typically, one arm stabilizes a movable object while the other performs the insertion. 

\textbf{Articulated Object Interaction.}
As in the \singlearm{} case, the robot must manipulate articulated objects; however now, two objects are manipulated simultaneously. 

\textbf{Long-Horizon Manipulation.}
Similar to the \singlearm{} case, these tasks involve executing sequential manipulation primitives. 

\subsection{Perturbations}
\label{method-pf}

Beyond standard in-distribution evaluation, the \benchmark{} measures robustness under a total of \numperturbationstotal{} visual, linguistic, physical, and ecological perturbations to assess out-of-distribution generalization. For all perturbations, we adopt the same terminology from the Colosseum \cite{pumacay2024colosseum}, where the Manipulation Object (MO) is the object that is directly manipulated, whereas the Receiving Object (RO) is the object that is indirectly interacted with.

\textbf{Visual Perturbations.}
These modifications alter scene appearance without changing the task's language command or physical configuration. The visual perturbations include changes to object colors (\mocolor{}, \rocolor{}), object textures (\motexture{}, \rotexture{}), lighting conditions (\lightcolor{}), table appearance (\tablecolor{}, \tabletexture{}), background appearance (\backgroundcolor{}, \backgroundtexture{}), distractor objects (\distractorobjects{}), and camera pose (\camerapose{}).

\textbf{Language Perturbations.}
Language robustness is evaluated under four perturbation conditions. \texttt{Language-Paraphrase} uses a semantically equivalent instruction, \texttt{Language-Other-Task} replaces it with an instruction from a different task, \texttt{Language-Random} uses random English words, and \texttt{Language-None} removes the instruction entirely. Together, these perturbations evaluate robustness to increasingly corrupted language inputs.

\textbf{Action Perturbations.}
Action perturbations test robustness to changes in object properties and initial conditions. These include manipulation-object size (\mosize{}), receiving-object size (\rosize{}), and randomized initial poses (\poserandomization{}).

\textbf{Combined Perturbations.}
Combinations of perturbations represent the most challenging conditions a model may face. To evaluate this, the \allperturbations{} condition applies all perturbations simultaneously.



\subsection{Demonstrations}

Demonstrations are generated by expert motion planning solvers for each task, with privileged simulation state available during planning. Each timestep in a demonstration contains the robot's state, an action, and an observation. The robot's state consists of joint positions and velocities. The action space is absolute pose control and absolute joint control for the \singlearm{} and \bimanual{} test suites, respectively. All task environments include 2 external RGB cameras, $\mathcal{O}_\textrm{rgb-ext} \in \mathbb{R}^{224 \times 224 \times 3}$, and a wrist-mounted RGB camera on each arm, $\mathcal{O}_\textrm{rgb-wrist} \in \mathbb{R}^{128 \times 128 \times 3}$. Thus, the observation space for \singlearm{} tasks is $\mathcal{O}_\textrm{rgb-ext} \times \mathcal{O}_\textrm{rgb-ext} \times \mathcal{O}_\textrm{rgb-wrist}$, and for \bimanual{} tasks is $\mathcal{O}_\textrm{rgb-ext} \times \mathcal{O}_\textrm{rgb-ext} \times \mathcal{O}_\textrm{rgb-wrist} \times \mathcal{O}_\textrm{rgb-wrist}$.

%
\begin{figure*}[t]
\topfigureslidedown
    \centering
    \includegraphics[width=0.98\textwidth]{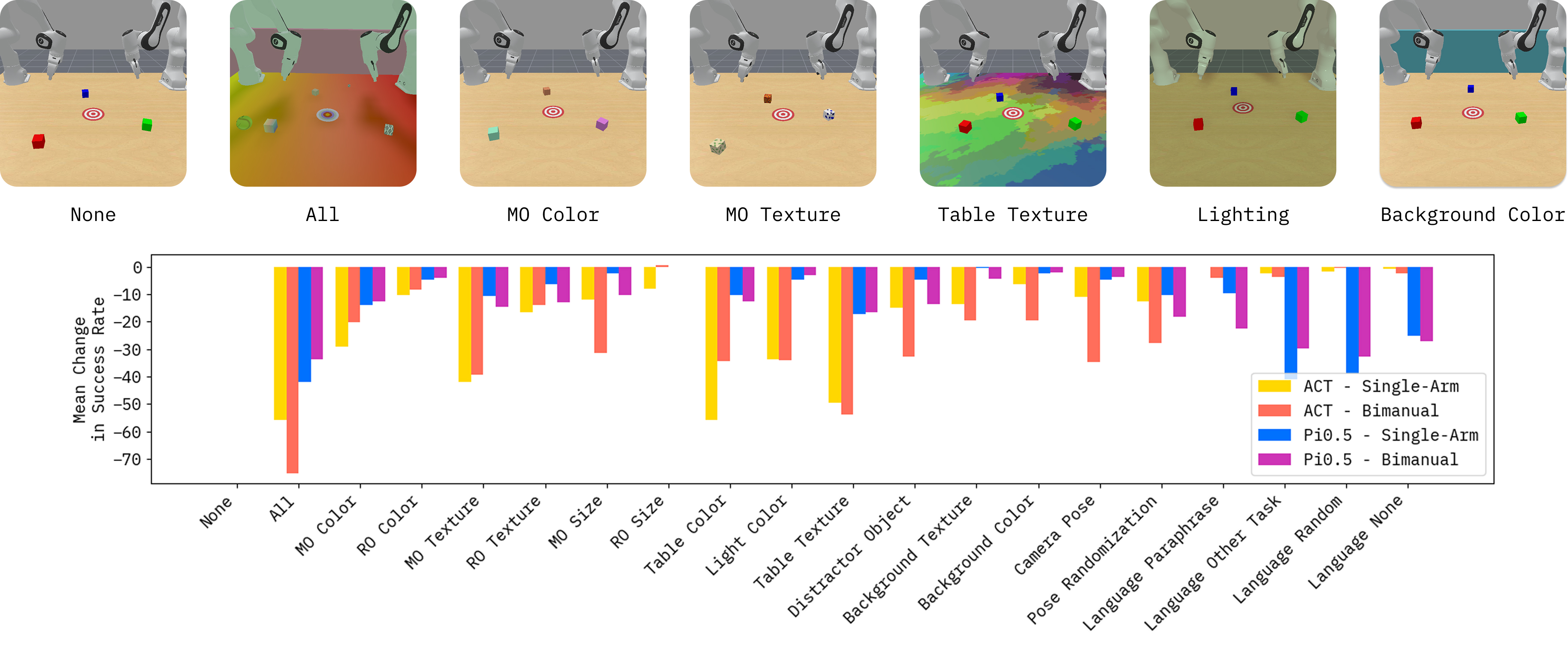}
    \caption{Average change in unnormalized success rate for each perturbation. The top row illustrates select perturbations for the \DualArmStackTwoCubes{} task. The average is calculated from only environments with a base success rate (no perturbations) of at least 10\% to reduce the noise in the aggregate metrics. The raw results as well as a list of which tasks are included in the perturbation averages are included in the Appendix, which is available on the project's website: \website.
    }
    \label{fig:mean_percent_change}
\end{figure*}

%
\begin{figure}
    \centering
    \includegraphics[width=1.0\columnwidth]{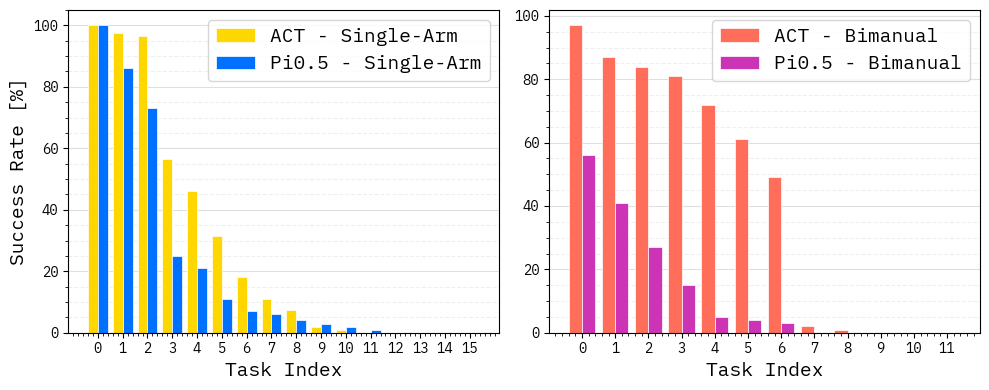}
    \caption{
      Distribution of per-task no-perturbation success rates of all models. 
      The x-axis orders tasks by success rate; consequently, the ordering between models may differ, so the $i$-th success rate does not necessarily correspond to the same environment between different models. 
      \act{} outperforms \piO{} on both test suites.}
    \label{fig:waterfall}
\end{figure}

\section{Experimental Setup}


We provide evaluations of two state-of-the-art language-conditioned imitation learning models: \act{} and \piO{} \cite{act-aloha,intelligence2025pi05visionlanguageactionmodelopenworld}. A separate multitask policy is trained for each model and robot morphology, yielding four policies total.

\subsection{Implementations}

The \piO{} implementation is provided by the LeRobot GitHub repository\cite{cadene2024lerobot}. The \act{} implementation is a modified version from ManiSkill\cite{taomaniskill3} which incorporates a language embedding input to the model. The language embeddings are generated from the pretrained open-source frozen CLIP language-side encoder\cite{radford2021learningtransferablevisualmodels}. Instructions for training and running both models are available on the project's website: \website.

\subsection{Training and Evaluation}
We train each model on a dataset composed of 100 perturbation-free demonstrations per task. Next, following prior work~\cite{pumacay2024colosseum}, each is evaluated on every perturbation individually. To ensure robust statistical estimates, success rates are measured over 200 episodes for every perturbation. All models were trained on a single NVIDIA H100 GPU. The ACT \singlearm{} and \bimanual{} models were trained with a batch size of 256 and a learning rate of 1e-4 until convergence, which corresponds to 130k iterations and 230k iterations for the \singlearm{} and \bimanual{} models, respectively. The \piO{} models were trained with a batch size of 4 and a learning rate of 1e-5, corresponding to 500k and 250k iterations in the \singlearm{} and \bimanual{} models, respectively.

\subsection{Hardware Validation}
\begin{figure}[t]
\topfigureslidedown
    \centering
    \includegraphics[width=0.98\columnwidth]{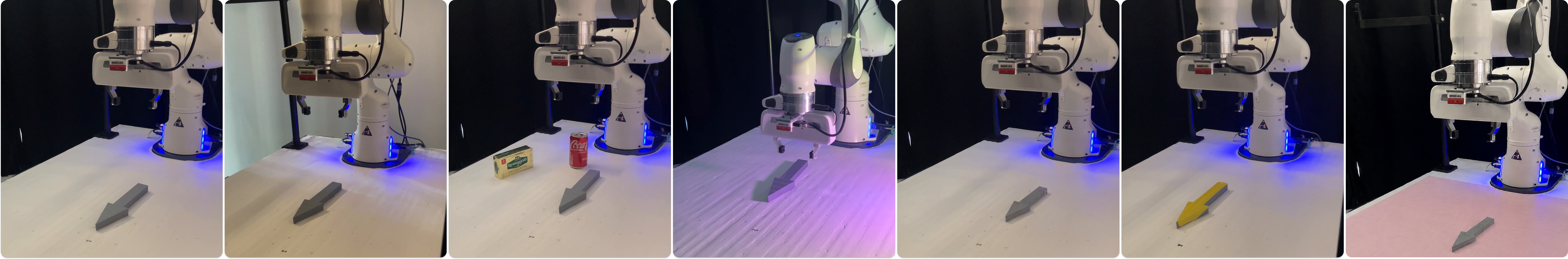}
    \caption{
    Hardware setup for \RotateArrow{}.
    From left to right, the perturbations are \none{} / \languagenone{}, \backgroundcolor{}, \distractorobjects{}, \lightcolor{}, \mosize{}, \mocolor{}, and \tablecolor{}.
    Additional hardware tasks are shown in the Appendix which is included in the project's website: \website}
    \label{fig:hardware_setup}
\end{figure}

To assess whether simulation captures real-world robustness trends, we reproduce five \singlearm{} PJG tasks in the real world: \RaiseCube{}, \RotateArrow{}, \LiftPegUpright{}, \PickDishFromRack{}, and \PickSodaFromCabinet{} and train a single \act{} multitask model from scratch using 60 demonstrations per task. Each task is evaluated under each of the \none{}, \mosize{}, \mocolor{}, \backgroundcolor{}, \lightcolor{}, \languagenone{}, and \tablecolor{} perturbations across 20 trials per task-condition pair. Images of each perturbation are presented in Figure \ref{fig:hardware_setup}.

\section{Results and Analysis}
\label{sec:results}
In this section, the following questions are addressed: How do the models generalize along the vision, language, and action axes? How do absolute success rates compare between models? How does embodiment affect performance? How well do simulation results predict performance on hardware?

\subsection{Vision Generalization}




\act{} exhibits worse visual generalization compared to \piO{}. Numerically, the average decrease in success rate for \act{} is 28.57\% and 29.06\% on the \singlearm{} and \bimanual{} test suites respectively, whereas it is only 7.73\% and 9.86\% for \piO{}. 
We hypothesize this difference is driven by the scale of pretraining on the base \piO{} model and the size of the vision encoder.
Specifically, \piO{} uses a $\sim$400M parameter SigLIP~\cite{zhai2023sigmoidlosslanguageimage} model, whereas \act{} uses a ResNet18~\cite{7780459} model with $\sim$11M parameters. 
While the exact size of the pretraining dataset used by \piO{} is not specified, there are 280k gradient steps taken on multimodal web data, likely providing a strong visual prior. 
In contrast, the \act{} model is not pretrained. 
Prior  work\cite{lee2025molmoactactionreasoningmodels,intelligence2025pi05visionlanguageactionmodelopenworld,brohan2023rt2,kim24openvla,jiang2023vimageneralrobotmanipulation} suggests pretraining a policy on multimodal web data (e.g. image captioning, visual question answering, object localization, and point labeling) improves visual robustness. 



%

\subsection{Language Generalization}

Surprisingly, \act{} exhibits a nearly negligible performance drop on all four language perturbations. Under \languageparaphrase{} for example, the average percent change in success rate is -0.28\% and -0.57\% for \singlearm{} and \bimanual{}, respectively, as opposed to -19.16\% and -24.0\% for \piO{}. We hypothesize that \act{} simply ignores the language conditioning in its input.


For \piO{}, the four language perturbations reveal a clear relationship between the severity of the language corruption and task success rate (Figure~\ref{fig:mean_percent_change}): as the language input deviates further from the training distribution, performance degrades more substantially. For the bimanual \piO{} model, the average relative change in success rate compared to the no-perturbation condition is $-63.6\%$, $-72.2\%$, $-87.8\%$, and $-91.2\%$ for \languageparaphrase{}, \languagenone{}, \languageothertask{}, and \languagerandom{}, respectively across all tasks. This progression is consistent with the intuition that semantically equivalent paraphrases remain closest to the training distribution, whereas unrelated task instructions and random English words represent increasingly severe distribution shifts.

\begin{figure}[t]
    \centering
    \includegraphics[width=0.98\columnwidth]{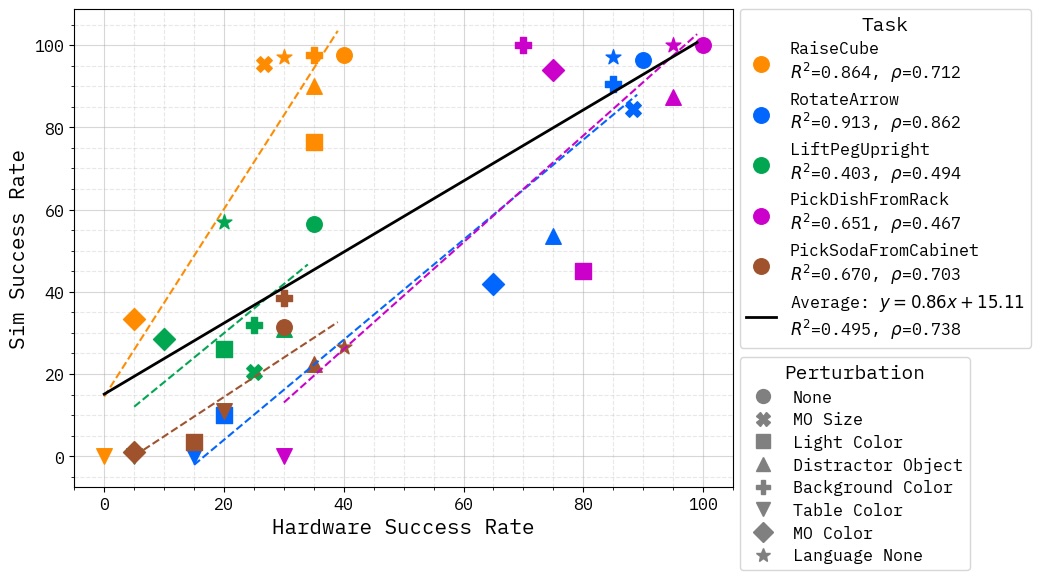}
    \caption{
            Simulation vs. hardware success rate comparison across five tasks. While simulation does not perfectly predict the success rate achieved on hardware, it does capture the overall trends across perturbations.
        }

    \label{fig:hardware_vs_sim}
\end{figure}

%
\begin{figure}[t]
    \centering
    \includegraphics[width=0.98\columnwidth]{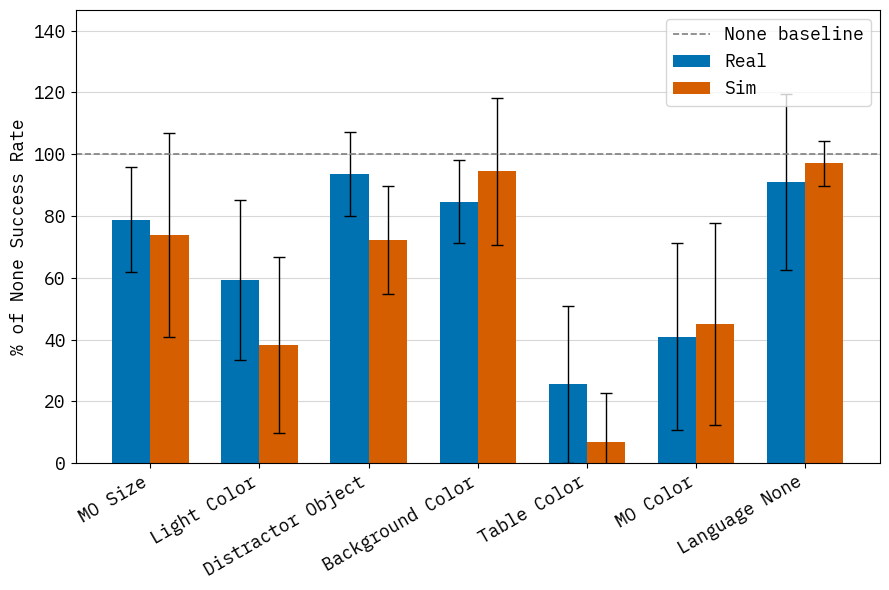}
    \caption{
            Relative robustness under each perturbation, normalized by the corresponding no-perturbation success rate. Bars report the average normalized success rate across the five hardware tasks for both simulation and real-world experiments. Error bars denote one standard deviation across tasks.
    }

    \label{fig:hardware_vs_sim_relative_change}
\end{figure}

\subsection{Action Generalization}



The average relative decrease in task success rate across the action perturbations is 11.2\% and 18.24\% (\singlearm{}, \bimanual{}) for \act{} compared to 10.08\% and 4.11\% for \piO{}. The primary contributor for this performance degradation is the \poserandomization{} perturbation. Under this perturbation, the average decrease in success rate is 21.8\% and 40.9\% (\singlearm{}, \bimanual{}) for \act{}, and 12.9\% and 51.4\% for \piO{}. Notably, \poserandomization{} results in a greater performance drop in the \bimanual{} tasks compared to the \singlearm{} tasks. We hypothesize this performance drop is due to increased spatial reasoning demands arising from the larger action space of the \bimanual{} morphology. 
Additionally, the use of joint space control for the \bimanual{} tasks as opposed to 6-DOF cartesian pose control used in the \singlearm{} test suite may contribute to the performance drop. 
In contrast to \poserandomization{}, the \mosize{} perturbation has a neutral or positive impact on success rate, as increasing the size of the MO object often makes a task easier. 
For example, in \ScoopBanana{} the MO object (a banana) is more easily scooped when its size is increased, which reduces the task's difficulty. Overall, \poserandomization{} remains a primary challenge for both models.



\subsection{Absolute Success Rate Comparison}

Between the two models, \act{} achieves higher success rates across all tasks with no perturbations compared to \piO{} (Figure \ref{fig:waterfall}). 
Quantitatively, the mean of the success rates across all tasks with no perturbations is 29.2\% and 44.5\% for \act{} on the \singlearm{} and \bimanual{} test suites, respectively, compared to 21.2\% and 12.6\% for \piO{}, corresponding to increases of $1.37\times$ and $3.53\times$.
One possible explanation is that \act{} overfits to the \benchmark{} tasks as it is trained exclusively on demonstrations from the \benchmark{}. In contrast, \piO{} maintains generally good behavior on a wide range of tasks at the expense of in-distribution performance. In other words, \piO{} is a better generalist but worse specialist.
Consistent with this interpretation, \piO{} demonstrates improved visual and action generalization (Figure~\ref{fig:mean_percent_change}), highlighting a tradeoff between base task proficiency and robustness to changes in the environment.
Another possibility is the larger model capacity of \piO{} requires more fine-tuning data than is available, whereas the dataset size is sufficient for \act{} given its smaller model size. 
Lastly, we note that the training setups differ in terms of batch size, learning rate, and number of updates, which may contribute to the observed gap; however, these configurations follow standard practices for each model family. 
These results support the claim that strong base performance does not imply strong generalization across visual, language, or action axes.


\subsection{Differences between Embodiments}


A remarkably similar decrease in success rate is observed across morphologies between the two models. For example, the average change in `Mean Absolute Change in Success Rate' between a perturbation over the \singlearm{} and \bimanual{} test suites is only 10.83\%. For \piO{}, the difference is even lower at 3.69\%. These values indicate that the perturbations reliably determine a model's intrinsic generalization, as their effects remain consistent across morphologies, action spaces, and tasks.




\subsection{Hardware Validation}

Real-world experiments indicate that \benchmark{} does not precisely predict the \textit{absolute} success rate achieved on hardware due to the sim-to-real gap. This is visible in Figure~\ref{fig:hardware_vs_sim}, where the coefficient of determination across all 35 task--perturbation pairs is $R^2 = 0.49$. However, the best-fit relationship between all simulation and hardware success rates, $\text{SR}_{\mathrm{sim}} = 0.86 * \text{SR}_{\mathrm{real}} + 15.1$, indicates that simulation largely tracks changes in hardware performance despite the sim-to-real gap. While there remains room for improvement in predicting absolute success rates, the \textit{relative} effects of perturbations with respect to the no-perturbation baseline \textit{are} well preserved between simulation and hardware (Figure~\ref{fig:hardware_vs_sim_relative_change}). Numerically, after normalizing each perturbation by the corresponding no-perturbation success rate, the mean absolute difference between simulation and hardware is only $12.3$ percentage points, suggesting that \benchmark{} provides a useful estimate of the \emph{relative} robustness of manipulation policies under distribution shifts despite differences in absolute task success. We note that our hardware validation is limited to a single policy, five manipulation tasks, and seven perturbation conditions. While these experiments provide encouraging evidence for the sim-to-real validity of \benchmark{}, broader hardware evaluation across additional policies, tasks, and perturbation types would further strengthen these conclusions.



%
%

\section{Conclusion}

In this work, we introduce \benchmark{}, a large-scale, GPU-parallelized benchmark designed to systematically evaluate generalization in VLA models along visual, language, and action axes. 
Our results show that \act{} appears to rely weakly on language input, whereas \piO{} remains comparatively robust to visual perturbations. Both models remain sensitive to changes in initial object poses, particularly in the Bimanual test suites. 

Furthermore, simulation results were shown to predict relative performance changes in the real world, supporting the utility of the proposed benchmark as a proxy for real-world evaluation despite remaining sim-to-real gaps. 
However, \benchmark{} remains limited by rigid-body simulation and does not capture deformable-object manipulation or the full complexity of contact-rich effects such as frictional uncertainty, compliance, force feedback, and stick--slip transitions. Extending the benchmark to include these effects represents important future work.
By standardizing tasks, perturbation factors, and evaluation protocols, \benchmark{} provides a scalable and reproducible framework for benchmarking generalization in robot learning. 
We expect this benchmark to facilitate rigorous analysis of generalization and to drive the development of models that maintain high success rates under diverse visual, language, and action perturbations.

\bibliographystyle{IEEEtran} 
\vspace{-4pt}
\bibliography{references}  

@inproceedings{pumacay2024colosseum,
  author    = {Pumacay, Wilbert and Singh, Ishika and Duan, Jiafei and Krishna, Ranjay and Thomason, Jesse and Fox, Dieter},
    title     = {THE COLOSSEUM: A Benchmark for Evaluating Generalization for Robotic Manipulation},
    BOOKTITLE = {Proceedings of Robotics: Science and Systems}, 
    YEAR      = {2024}, 
}

@misc{zhao2025manipbenchbenchmarkingvisionlanguagemodels,
      title={ManipBench: Benchmarking Vision-Language Models for Low-Level Robot Manipulation}, 
      author={Enyu Zhao and Vedant Raval and Hejia Zhang and Jiageng Mao and Zeyu Shangguan and Stefanos Nikolaidis and Yue Wang and Daniel Seita},
      year={2025},
      eprint={2505.09698},
      archivePrefix={arXiv},
      primaryClass={cs.RO},
      url={https://arxiv.org/abs/2505.09698}, 
}

@misc{zheng2022vlmbenchcompositionalbenchmarkvisionandlanguage,
      title={VLMbench: A Compositional Benchmark for Vision-and-Language Manipulation}, 
      author={Kaizhi Zheng and Xiaotong Chen and Odest Chadwicke Jenkins and Xin Eric Wang},
      year={2022},
      eprint={2206.08522},
      archivePrefix={arXiv},
      primaryClass={cs.RO},
      url={https://arxiv.org/abs/2206.08522}, 
}

@misc{zhang2024vlabenchlargescalebenchmarklanguageconditioned,
      title={VLABench: A Large-Scale Benchmark for Language-Conditioned Robotics Manipulation with Long-Horizon Reasoning Tasks}, 
      author={Shiduo Zhang and Zhe Xu and Peiju Liu and Xiaopeng Yu and Yuan Li and Qinghui Gao and Zhaoye Fei and Zhangyue Yin and Zuxuan Wu and Yu-Gang Jiang and Xipeng Qiu},
      year={2024},
      eprint={2412.18194},
      archivePrefix={arXiv},
      primaryClass={cs.RO},
      url={https://arxiv.org/abs/2412.18194}, 
}

@article{mees2022calvin,
author = {Oier Mees and Lukas Hermann and Erick Rosete-Beas and Wolfram Burgard},
title = {CALVIN: A Benchmark for Language-Conditioned Policy Learning for Long-Horizon Robot Manipulation Tasks},
journal={IEEE Robotics and Automation Letters (RA-L)},
volume={7},
number={3},
pages={7327-7334},
year={2022}
}

@misc{kim2026liberoparadiagnosticbenchmarkmetrics,
      title={LIBERO-Para: A Diagnostic Benchmark and Metrics for Paraphrase Robustness in VLA Models}, 
      author={Chanyoung Kim and Minwoo Kim and Minseok Kang and Hyunwoo Kim and Dahuin Jung},
      year={2026},
      eprint={2603.28301},
      archivePrefix={arXiv},
      primaryClass={cs.LG},
      url={https://arxiv.org/abs/2603.28301}, 
}

@INPROCEEDINGS{7780459,
  author={He, Kaiming and Zhang, Xiangyu and Ren, Shaoqing and Sun, Jian},
  booktitle={2016 IEEE Conference on Computer Vision and Pattern Recognition (CVPR)}, 
  title={Deep Residual Learning for Image Recognition}, 
  year={2016},
  volume={},
  number={},
  pages={770-778},
  doi={10.1109/CVPR.2016.90}}

@misc{zhai2023sigmoidlosslanguageimage,
      title={Sigmoid Loss for Language Image Pre-Training}, 
      author={Xiaohua Zhai and Basil Mustafa and Alexander Kolesnikov and Lucas Beyer},
      year={2023},
      eprint={2303.15343},
      archivePrefix={arXiv},
      primaryClass={cs.CV},
      url={https://arxiv.org/abs/2303.15343}, 
}

@article{ma2022vip,
  title={VIP: Towards Universal Visual Reward and Representation via Value-Implicit Pre-Training},
  author={Ma, Yecheng Jason and Sodhani, Shagun and Jayaraman, Dinesh and Bastani, Osbert and Kumar, Vikash and Zhang, Amy},
  journal={arXiv preprint arXiv:2210.00030},
  year={2022}
}

@misc{jiang2023vimageneralrobotmanipulation,
      title={VIMA: General Robot Manipulation with Multimodal Prompts}, 
      author={Yunfan Jiang and Agrim Gupta and Zichen Zhang and Guanzhi Wang and Yongqiang Dou and Yanjun Chen and Li Fei-Fei and Anima Anandkumar and Yuke Zhu and Linxi Fan},
      year={2023},
      eprint={2210.03094},
      archivePrefix={arXiv},
      primaryClass={cs.RO},
      url={https://arxiv.org/abs/2210.03094}, 
}

@misc{lee2025molmoactactionreasoningmodels,
      title={MolmoAct: Action Reasoning Models that can Reason in Space}, 
      author={Jason Lee and Jiafei Duan and Haoquan Fang and Yuquan Deng and Shuo Liu and Boyang Li and Bohan Fang and Jieyu Zhang and Yi Ru Wang and Sangho Lee and Winson Han and Wilbert Pumacay and Angelica Wu and Rose Hendrix and Karen Farley and Eli VanderBilt and Ali Farhadi and Dieter Fox and Ranjay Krishna},
      year={2025},
      eprint={2508.07917},
      archivePrefix={arXiv},
      primaryClass={cs.RO},
      url={https://arxiv.org/abs/2508.07917}, 
}

@inproceedings{james2019rlbenchrobotlearningbenchmark,
  title={RLBench: The Robot Learning Benchmark and Learning Environment},
  author={James, Stephen and Davison, Andrew J. and Johns, Edward},
  booktitle={IEEE Robotics and Automation Letters},
  year={2019}
}

@inproceedings{james2019pyrepbringingvrepdeep,
  title={PyRep: Bringing V-REP to Deep Robot Learning},
  author={James, Stephen and Ma, Zicong and Arrojo, Diego Rodriguez and Davison, Andrew J.},
  booktitle={Conference on Robot Learning (CoRL)},
  year={2019}
}

@misc{coppeliaSim,
  title={CoppeliaSim Robot Simulator},
  author={{Coppelia Robotics}},
  year={2022},
  note={https://www.coppeliarobotics.com}
}

@inproceedings{murali2020roboverse,
  title={RoboVerse: Towards a Unified Platform for Robotic Manipulation},
  author={Murali, Adithyavairavan and others},
  booktitle={Conference on Robot Learning Workshop},
  year={2020}
}

@misc{robotwin,
  title={RobotWin: A Platform for Scalable Robot Learning},
  author={RobotWin Team},
  year={2024},
  note={https://robotwin-platform.github.io}
}

@misc{roboarena,
  title={RoboArena: Distributed Real-World Evaluation of Generalist Robot Policies},
  author={RoboArena Team},
  year={2024}
}

@misc{bimanual,
  title={Bimanual Manipulation Benchmark},
  author={Bimanual Benchmark Team},
  year={2024},
  note={https://bimanual.github.io}
}

@inproceedings{r3m,
  title={R3M: A Universal Visual Representation for Robot Manipulation},
  author={Nair, Suraj and others},
  booktitle={Conference on Robot Learning},
  year={2022}
}

@inproceedings{mvp,
  title={MVP: Multi-View Pretraining for Vision-Language Robotics},
  author={Xiao, Tete and others},
  booktitle={Conference on Robot Learning},
  year={2022}
}

@inproceedings{cliport,
  title={CLIPort: What and Where Pathways for Robotic Manipulation},
  author={Shridhar, Mohit and others},
  booktitle={Conference on Robot Learning},
  year={2022}
}

@inproceedings{huang2023voxposer,
  title={VoxPoser: Composable 3D Value Maps for Robotic Manipulation with Language Models},
  author={Huang, Wenlong and others},
  booktitle={Conference on Robot Learning},
  year={2023}
}

@article{brohan2023rt2,
  title={RT-2: Vision-Language-Action Models Transfer Web Knowledge to Robotic Control},
  author={Brohan, Anthony and others},
  journal={arXiv preprint arXiv:2307.15818},
  year={2023}
}

@misc{padalkar2023open,
  title={Open X-Embodiment: Robotic Learning Datasets and RT-X Models},
  author={Padalkar, Abhishek and others},
  year={2023},
  journal={arXiv preprint arXiv:2310.08864}
}

@article{kim24openvla,
  title={OpenVLA: Vision-Language-Action Models for Robotics},
  author={Kim, Moo Jin and others},
  journal={arXiv preprint arXiv:2406.09246},
  year={2024}
}

@article{driess2023palm,
  title={PaLM-E: An Embodied Multimodal Language Model},
  author={Driess, Danny and others},
  journal={arXiv preprint arXiv:2303.03378},
  year={2023}
}

@article{black2024pi0,
  title={$\pi 0$: A Vision-Language-Action Model for General Robot Control},
  author={Black, Kevin and others},
  journal={arXiv preprint arXiv:2405.03854},
  year={2024}
}

@article{pertsch2025pi0fast,
  title={$\pi 0$-FAST: Fast Vision-Language-Action Models for Robotics},
  author={Pertsch, Karl and others},
  journal={arXiv preprint arXiv:2501.00000},
  year={2025}
}

@misc{intelligence2025pi05visionlanguageactionmodelopenworld,
  title         = {{$\pi_{0.5}$}: A Vision-Language-Action Model with Open-World Generalization},
  author        = {{Physical Intelligence} and others},
  year          = {2025},
  eprint        = {2504.16054},
  archivePrefix = {arXiv},
  primaryClass  = {cs.LG},
  url           = {https://arxiv.org/abs/2504.16054},
}

@inproceedings{shridhar2022peract,
  title={PerAct: Perceiver-Actor for 6-DoF Manipulation},
  author={Shridhar, Mohit and others},
  booktitle={Robotics: Science and Systems},
  year={2022}
}

@inproceedings{goyal2023rvt,
  title={RVT: Robotic Vision Transformer for Manipulation},
  author={Goyal, Ankit and others},
  booktitle={Conference on Robot Learning},
  year={2023}
}

@article{goyal2024rvt,
  title={RVT-2: Scaling Vision Transformers for Robot Manipulation},
  author={Goyal, Ankit and others},
  journal={arXiv preprint},
  year={2024}
}

@inproceedings{gervet2023act3d,
  title={Act3D: 3D Feature Fields for Manipulation Policies},
  author={Gervet, Theophile and others},
  booktitle={Conference on Robot Learning},
  year={2023}
}

@inproceedings{sundaresan2023kite,
  title={KITE: Keyframe Imitation for Task Execution},
  author={Sundaresan, Priya and others},
  booktitle={Conference on Robot Learning},
  year={2023}
}

@inproceedings{c2farm,
  title={C2FARM: Coarse-to-Fine Imitation Learning for Manipulation},
  author={James, Stephen and others},
  booktitle={Conference on Robot Learning},
  year={2022}
}

@article{act-aloha,
  title={Learning Fine-Grained Bimanual Manipulation with ACT},
  author={Zhao, Tony and others},
  journal={arXiv preprint},
  year={2023}
}

@misc{radford2021learningtransferablevisualmodels,
      title={Learning Transferable Visual Models From Natural Language Supervision}, 
      author={Alec Radford and Jong Wook Kim and Chris Hallacy and Aditya Ramesh and Gabriel Goh and Sandhini Agarwal and Girish Sastry and Amanda Askell and Pamela Mishkin and Jack Clark and Gretchen Krueger and Ilya Sutskever},
      year={2021},
      eprint={2103.00020},
      archivePrefix={arXiv},
      primaryClass={cs.CV},
      url={https://arxiv.org/abs/2103.00020}, 
}

@misc{cadene2024lerobot,
    author = {Cadene, Remi and Alibert, Simon and Soare, Alexander and Gallouedec, Quentin and Zouitine, Adil and Palma, Steven and Kooijmans, Pepijn and Aractingi, Michel and Shukor, Mustafa and Aubakirova, Dana and Russi, Martino and Capuano, Francesco and Pascal, Caroline and Choghari, Jade and Moss, Jess and Wolf, Thomas},
    title = {LeRobot: State-of-the-art Machine Learning for Real-World Robotics in Pytorch},
    howpublished = "\url{https://github.com/huggingface/lerobot}",
    year = {2024}
}

@article{taomaniskill3,
  title={ManiSkill3: GPU Parallelized Robotics Simulation and Rendering for Generalizable Embodied AI},
  author={Stone Tao and Fanbo Xiang and Arth Shukla and Yuzhe Qin and Xander Hinrichsen and Xiaodi Yuan and Chen Bao and Xinsong Lin and Yulin Liu and Tse-kai Chan and Yuan Gao and Xuanlin Li and Tongzhou Mu and Nan Xiao and Arnav Gurha and Viswesh Nagaswamy Rajesh and Yong Woo Choi and Yen-Ru Chen and Zhiao Huang and Roberto Calandra and Rui Chen and Shan Luo and Hao Su},
  journal = {Robotics: Science and Systems},
  year={2025},
}

@article{liu2024libero,
  title={Libero: Benchmarking knowledge transfer for lifelong robot learning},
  author={Liu, Bo and Zhu, Yifeng and Gao, Chongkai and Feng, Yihao and Liu, Qiang and Zhu, Yuke and Stone, Peter},
  journal={Advances in Neural Information Processing Systems},
  volume={36},
  year={2024}
}

@article{zhu2020robosuite,
  title     = {\href{https://arxiv.org/abs/2009.12293}{Robosuite: A modular simulation framework and benchmark for robot learning}},
  author    = {Zhu, Yuke and Wong, Josiah and Mandlekar, Ajay and Mart{\'\i}n-Mart{\'\i}n, Roberto and Joshi, Abhishek and
               Nasiriany, Soroush and Zhu, Yifeng},
  journal   = {arXiv preprint arXiv:2009.12293},
  year      = {2020},
  url       = {https://arxiv.org/abs/2009.12293}
}

@article{james2020rlbench,
  title     = {\href{https://arxiv.org/abs/1909.12271}{Rlbench: The robot learning benchmark \& learning environment}},
  author    = {James, Stephen and Ma, Zicong and Arrojo, David Rovick and Davison, Andrew J},
  journal   = {IEEE Robotics and Automation Letters},
  volume    = {5},
  number    = {2},
  pages     = {3019--3026},
  year      = {2020},
  publisher = {IEEE},
  url       = {https://ieeexplore.ieee.org/document/9001253}
}

@misc{openai2022chatgpt,
  title = {\text{ChatGPT}: Optimizing Language Models for Dialogue},
  author = {OpenAI},
  year = {2022},
  howpublished = {\url{https://openai.com/blog/chatgpt}},
  note = {Accessed: 2024-08-17}
}

@article{ravi2024sam2,
  title={\text{SAM 2}: Segment Anything in Images and Videos},
  author={Ravi, Nikhila and Gabeur, Valentin and Hu, Yuan-Ting and Hu, Ronghang and Ryali, Chaitanya and Ma, Tengyu and Khedr, Haitham and R{\"a}dle, Roman and Rolland, Chloe and Gustafson, Laura and Mintun, Eric and Pan, Junting and Alwala, Kalyan Vasudev and Carion, Nicolas and Wu, Chao-Yuan and Girshick, Ross and Doll{\'a}r, Piotr and Feichtenhofer, Christoph},
  journal={arXiv preprint arXiv:2408.00714},
  year={2024}
}

\end{document}